\documentclass[conference]{IEEEtran}
\IEEEoverridecommandlockouts
\usepackage{cite}
\usepackage{amsmath,amssymb,amsfonts}
\usepackage{algorithmic}
\usepackage{graphicx}
\usepackage{textcomp}
\usepackage{xcolor}
\usepackage{subcaption}
\usepackage{booktabs}
\usepackage{multirow}
 
\def\BibTeX{{\rm B\kern-.05em{\sc i\kern-.025em b}\kern-.08em
    T\kern-.1667em\lower.7ex\hbox{E}\kern-.125emX}}
\begin{document}

\title{Brno Mobile OCR Dataset
\thanks{This work has been supported by the Ministry of Culture Czech Republic in NAKI II project PERO (DG18P02OVV055) and by the Ministry of Education, Youth and Sports of the Czech Republic from the National Programme of Sustainability (NPU II), through the Project IT4Innovations Excellence in Science under Grant LQ1602.}
}

\author{\IEEEauthorblockN{Martin Ki\v{s}\v{s}}
\IEEEauthorblockA{\textit{Faculty of Information Technology} \\
\textit{Brno University of Technology}\\
Brno, Czech Republic \\
ikiss@fit.vutbr.cz}
\and
\IEEEauthorblockN{Michal Hradi\v{s}}
\IEEEauthorblockA{\textit{Faculty of Information Technology} \\
\textit{Brno University of Technology}\\
Brno, Czech Republic \\
ihradis@fit.vutbr.cz}
\and
\IEEEauthorblockN{Old\v{r}ich Kodym}
\IEEEauthorblockA{\textit{Faculty of Information Technology} \\
\textit{Brno University of Technology}\\
Brno, Czech Republic \\
ikodym@fit.vutbr.cz}
}

\maketitle

\begin{abstract}
We introduce the Brno Mobile OCR Dataset (B-MOD) for document Optical Character Recognition from low-quality images captured by handheld devices. 
While OCR of high-quality scanned documents is a mature field where many commercial tools are available, and large datasets of text in the wild exist,
no existing datasets can be used to develop and test document OCR methods robust to non-uniform lighting, image blur, strong noise, built-in denoising,
sharpening, compression and other artifacts present in many photographs from mobile devices.

This dataset contains 2~113 unique pages from random scientific papers, which were photographed by multiple people using 23~different mobile devices. 
The resulting 19~725 photographs of various visual quality are accompanied by precise positions and text annotations of 500k text lines. 
We further provide an evaluation methodology, including an evaluation server and a test set with non-public annotations. 

We provide a state-of-the-art text recognition baseline build on convolutional and recurrent neural networks trained with Connectionist Temporal Classification loss. 
This baseline achieves 2~\%, 22~\% and 73~\% word error rates on easy, medium and hard parts of the dataset, respectively, confirming that the dataset is challenging.

The presented dataset will enable future development and evaluation of document analysis for low-quality images. 
It is primarily intended for line-level text recognition, and can be further used for line localization, layout analysis, image restoration and text binarization.

\end{abstract}


\section{Introduction}

\begin{figure}[t]
  \centering
  \includegraphics[width=1\linewidth]{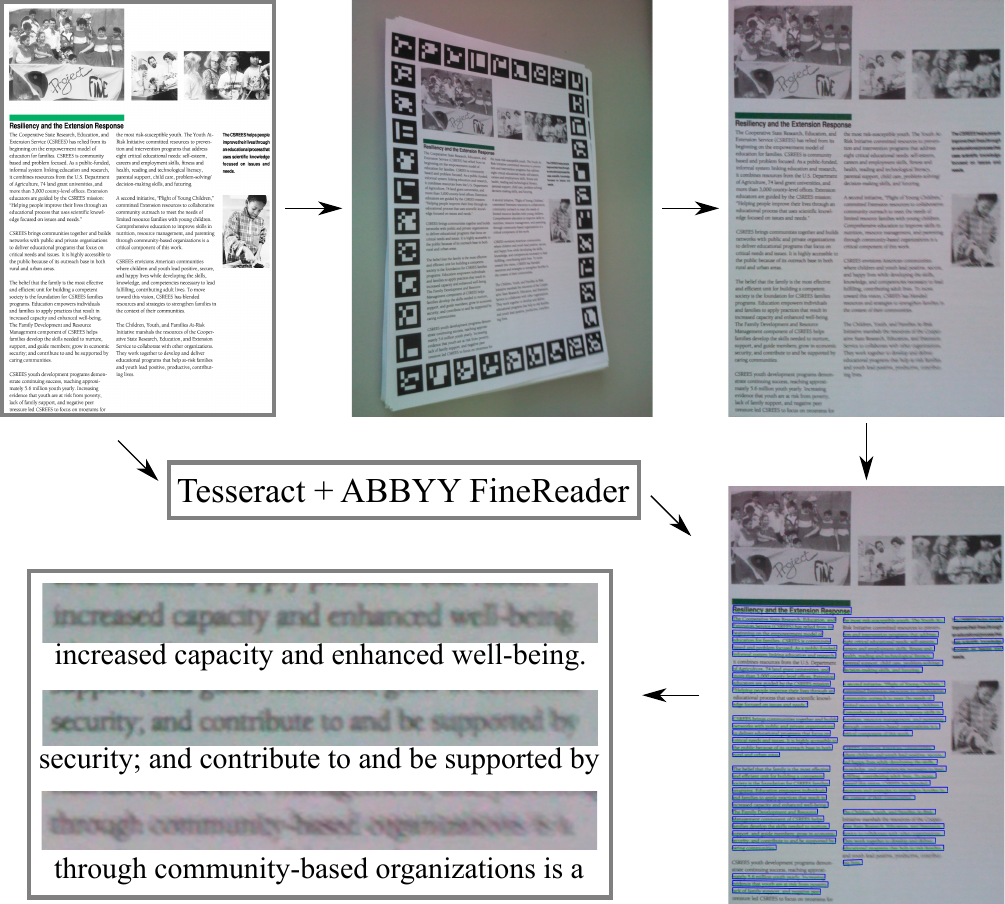}
  \caption{Dataset creation. Random pages are augmented by localization markers, printed, photographed, precisely localized, and matched with line annotations.}
  \label{fig:pipeline}
\end{figure}

Cameras in many mobile devices can capture images of high enough quality such that it is possible to use them to digitize whole printed document pages and to automatically transcribe their content. 
In fact, commercial applications providing exactly such functionality already exist\footnote{e.g. ABBYY FineScanner}, and can in some use-cases replace desktop scanners.
However, the OCR quality is often low due to poor illumination conditions, small print in large documents, or simply due to low quality of cameras in lower-end devices. 
Although printed text recognition methods which can handle significant image degradation such as low resolution, blur, noise, and non-uniform illumination, are studied in some application domains (e.g. car license plate recognition~\cite{Vasek2018}), such methods are not studied in the context of document digitization partially due to the fact that no suitable datasets to develop and test the methods exist.

We introduce the Brno Mobile OCR Dataset (B-MOD) which is specifically designed for the development and testing of document Optical Character Recognition methods from low-quality images that are often captured by handheld mobile devices. 
The dataset is unique due to the overall number of photographs, number of devices used, the amount of source documents used, and due to the fact that the photos were captured in unconstrained conditions by a large number of people. 
We believe that the dataset will enable future development and evaluation of document analysis methods for low-quality images.  

The following text reviews relevant document datasets and describes how the B-MOD dataset was created and its contents. 
The paper further introduces text recognition baselines and their results on the dataset, and finally conclusions are presented.

\section{Existing datasets}
\label{sec:existingDatasets}

Large number of datasets for document analysis and text recognition are already available. 
The datasets that are most closely related to B-MOD are SmartDoc~\cite{burie2015icdar2015}, SmartDoc-QA~\cite{nayef2015smartdoc} and SmartATID~\cite{chabchoub2016smartatid} which are intended for text recognition and printed document analysis captured by mobile devices. 
However, these datasets are relatively small, uniform and they lack line-level annotations.

\paragraph*{SmartDoc}
SmartDoc~\cite{burie2015icdar2015} is a dataset for English printed text recognition published in the Smartphone Document Capture and OCR Competition at ICDAR 2015. 
This dataset contains 12~100 photographs of 50 single column documents captured by two smartphones (Samsung Galaxy S4 and Nokia Lumia~920) in three different levels of illumination and forty predefined viewing positions. The smartphones were precisely positioned using a robotic arm and camera flash was not used. 
Some of the photos taken by Nokia Lumia~920 are slightly out of focus, otherwise the photographs are of high quality. 

The main disadvantages of this dataset are that it consists of small number of source documents and that it does not contain text line positions and corresponding line transcriptions, only page transcription is provided. 
Another limitation of the dataset is low variability of the lighting conditions and generally high quality of the photographs which make the dataset relatively simple for contemporary methods.

\paragraph*{SmartDoc-QA}
SmartDoc-QA \cite{nayef2015smartdoc} is similar to SmartDoc~\cite{burie2015icdar2015}, but it is primarily intended for Quality Assessment. 
This dataset contains 4260 photographs of 30 document pages, which are generally of lower quality compared to SmartDoc, including significant motion blur, out of focus blur, and non-uniform lighting. 
During acquisition, 5~different illumination conditions, 2~types of motion blur and 5~positions of smartphone with respect to the document were used.
Similar to SmartDoc, SmartDoc-QA is not suitable for training and evaluation of OCR systems due to the low number of source document pages and limited ground truth information.

\paragraph*{SmartATID}
SmartATID (Arabic Text Image Dataset) \cite{chabchoub2016smartatid} contains photographs of printed and handwritten Arabic documents captured using two smartphones (Samsung Galaxy S6 Edge and iPhone 6S Plus) which were attached to a tripod during acquisition. 
Illumination conditions and image degradations are similar as those in the SmartDoc-QA dataset. 
The dataset contains 25~560 images of 180~documents (116~printed and 64~handwritten documents). 
The dataset contains ground truth transcriptions of entire pages without any line information.

Many existing datasets focus on documents which are digitized using professional devices and workflow.
In such cases the challenge comes from variability, complexity, and ambiguity of the content. 
Most research and datasets focus on handwritten text documents. 
The IAM Handwriting Database~\cite{Marti2002} contains images of hand-transcribed English texts, the RIMES database~\cite{Augustin2006} contains filled forms and French letters, and The READ Dataset~\cite{sanchez2017icdar2017} contains historic letters.
The most challenging datasets for printed text focus on historic documents, e.g. the IMPACT dataset~\cite{papadopoulos2013impact}, the REID2017 dataset of Indian books~\cite{clausner2017icdar2017}, and The ENP Image and Ground Truth Dataset of Historical Newspapers~\cite{clausner2015enp}.

Another large group of datasets focuses on natural scene text recognition (e.g. Incidental Scene Text~\cite{karatzas2015icdar}, COCO-Text~\cite{gomez2017icdar2017} or DOST~\cite{iwamura2016downtown}).

\section{B-MOD Dataset}
\label{sec:ourDataset}

An overview of the process used to create the dataset is shown in Figure~\ref{fig:pipeline}. 
We photographed a large number of scientific article pages and precisely localized them using AR markers. 
The aligned photographs were supplemented by line-level ground truth from the source pages, the line position were refined, and the whole datasets was finally checked for any inconsistencies between the text ground truth and text line images.

\begin{figure}[t]
  \centering
  \includegraphics{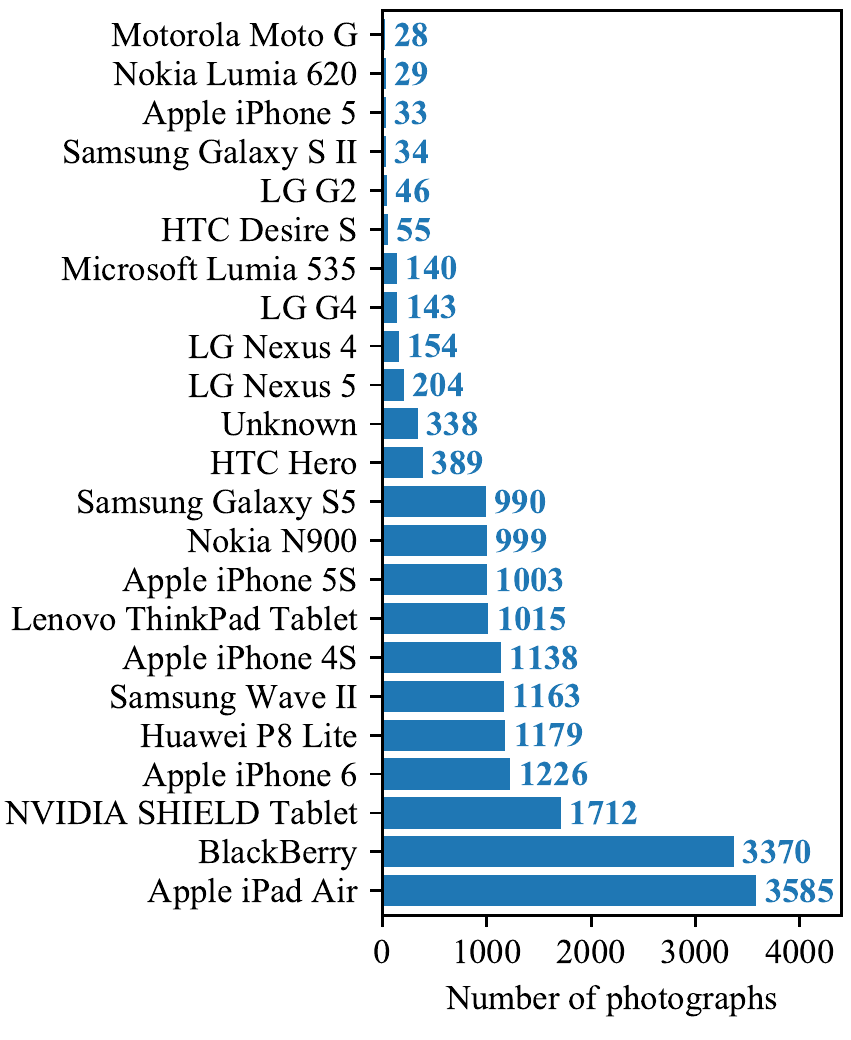}
  \caption{Histogram of used devices.}
  \label{fig:devices}
\end{figure}

\subsection{Data acquisition}
We collected 2113~pages \emph{templates} from random scientific articles downloaded from online preprint archives on various topics (information technology, medicine, chemistry, etc.). 
All of the collected pages are in English.

\begin{figure}
    \centering
    \begin{subfigure}{1\linewidth}
        \centering
        \includegraphics[width=0.45\linewidth]{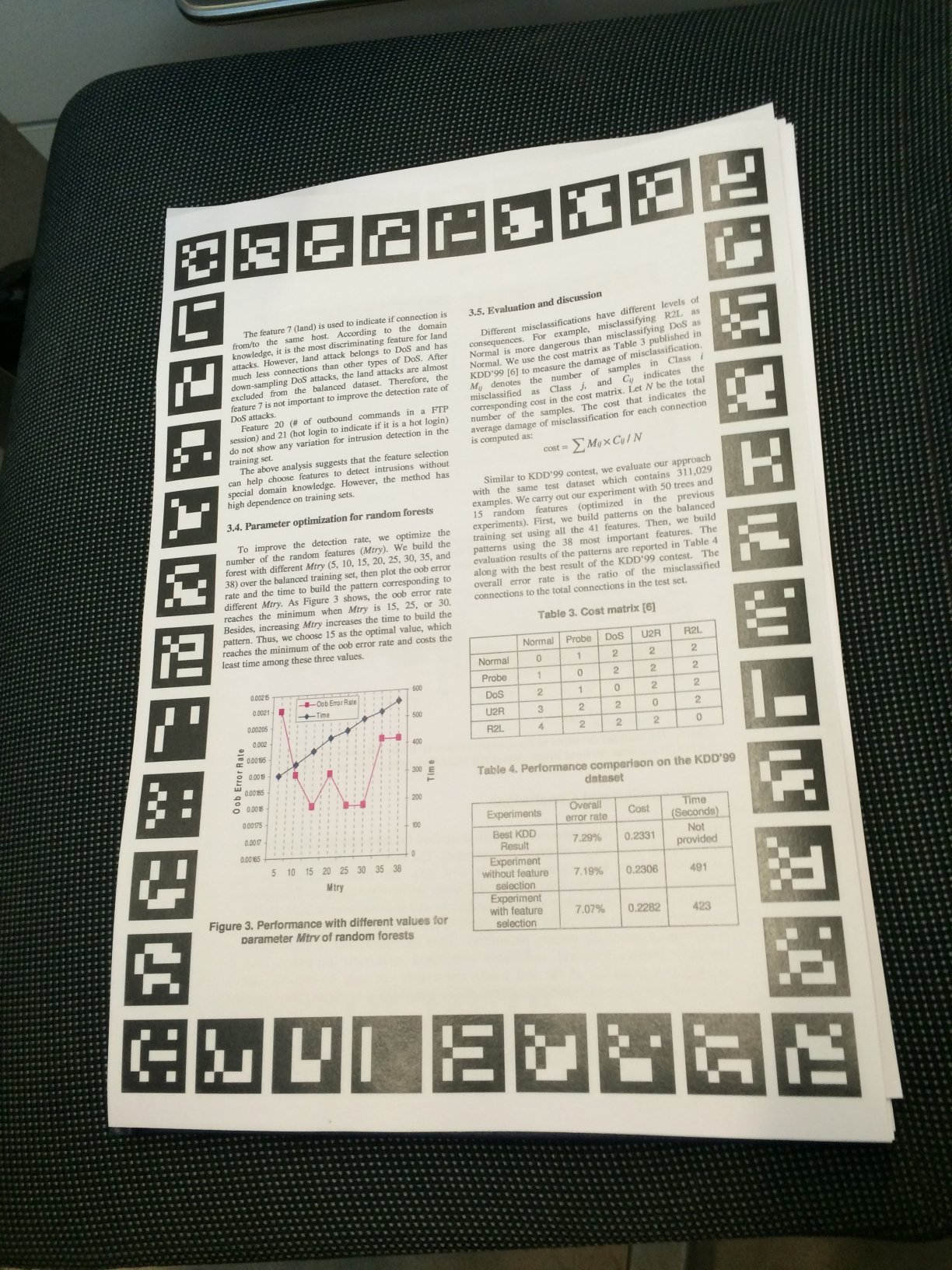}
        \hspace{0.1ex}
        \includegraphics[width=0.45\linewidth]{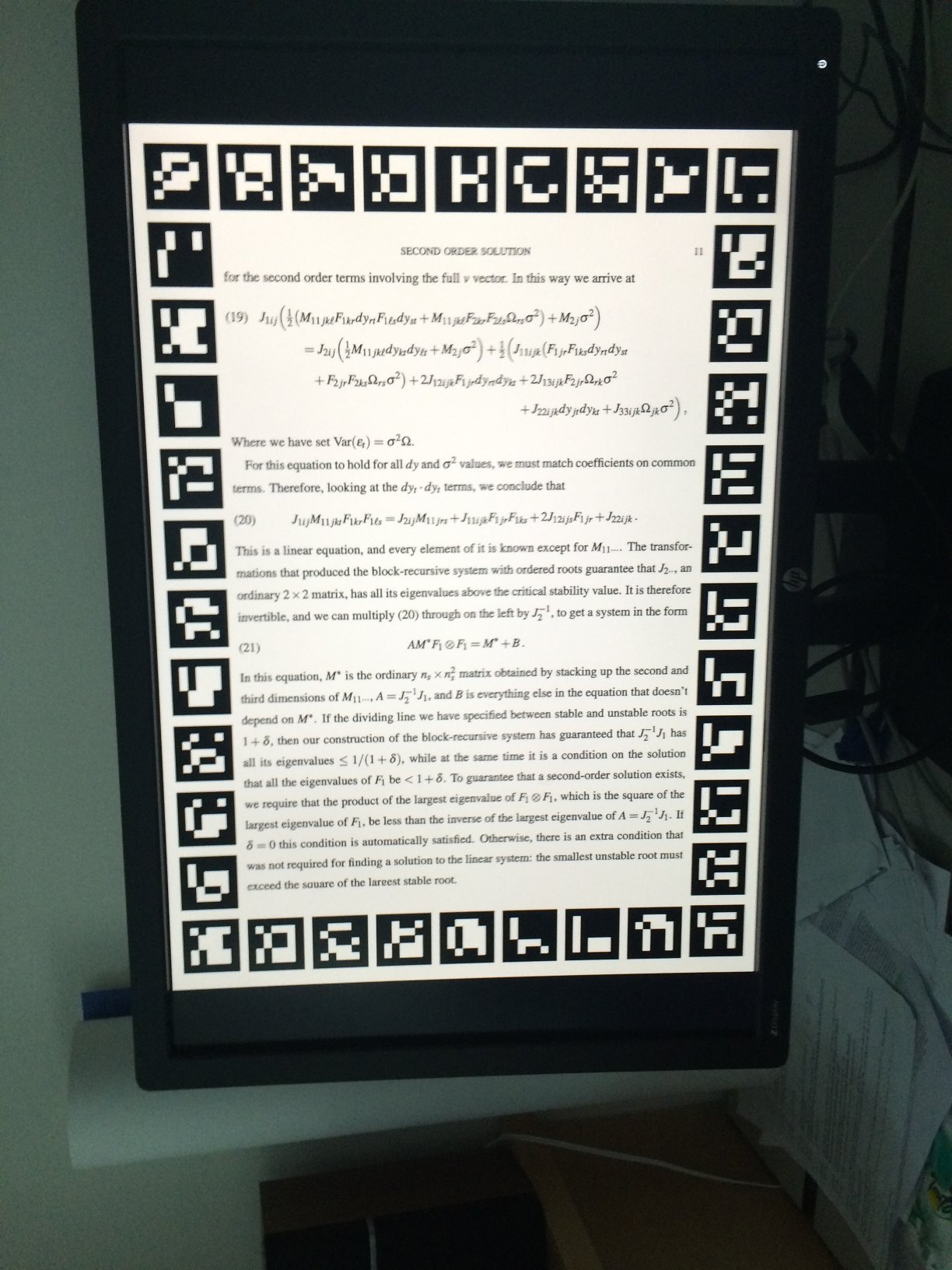}
    \end{subfigure}
    \par\medskip
    \begin{subfigure}{1\linewidth}
        \centering
        \includegraphics[width=0.45\linewidth]{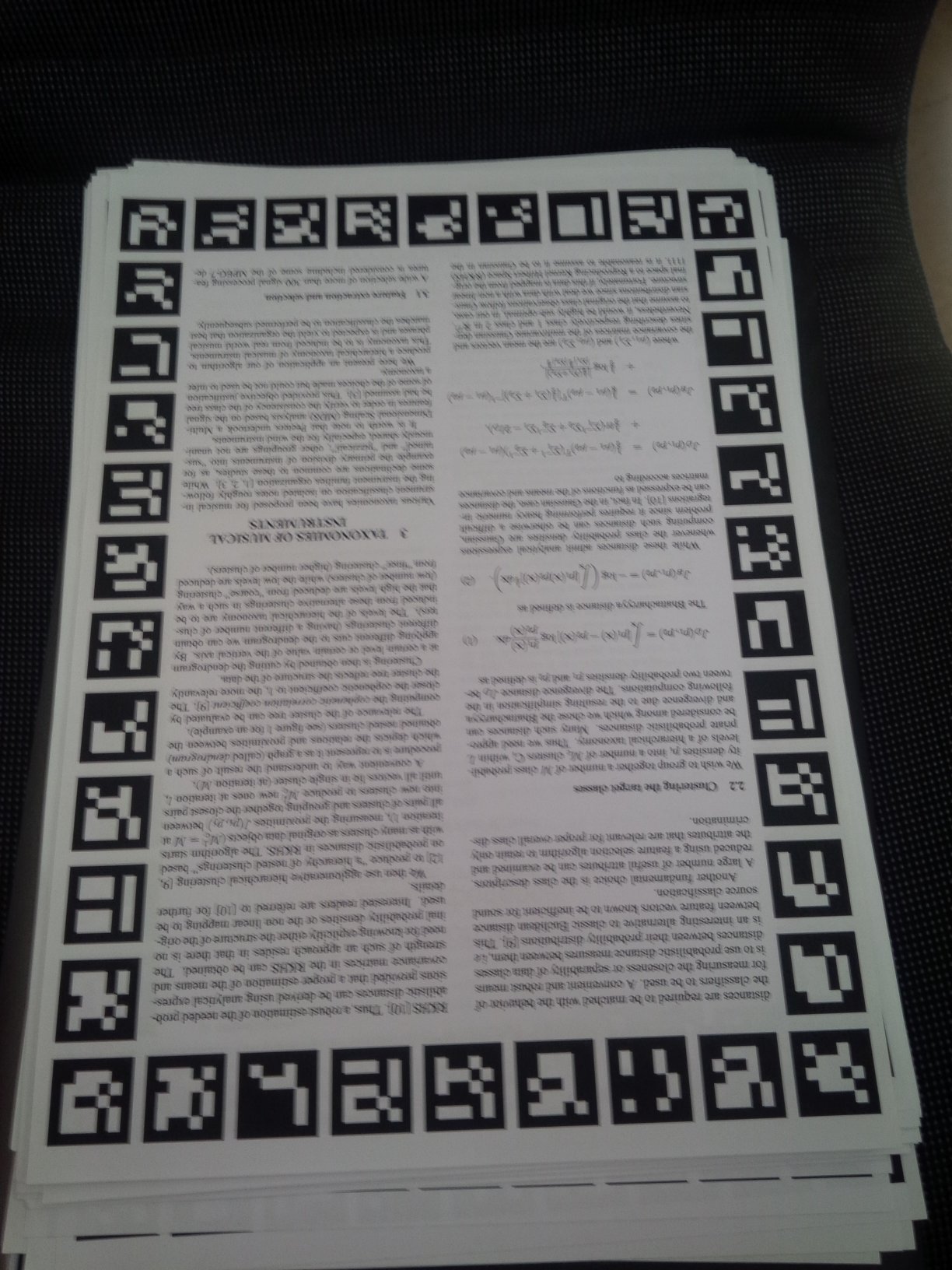}
        \hspace{0.1ex}
        \includegraphics[width=0.45\linewidth]{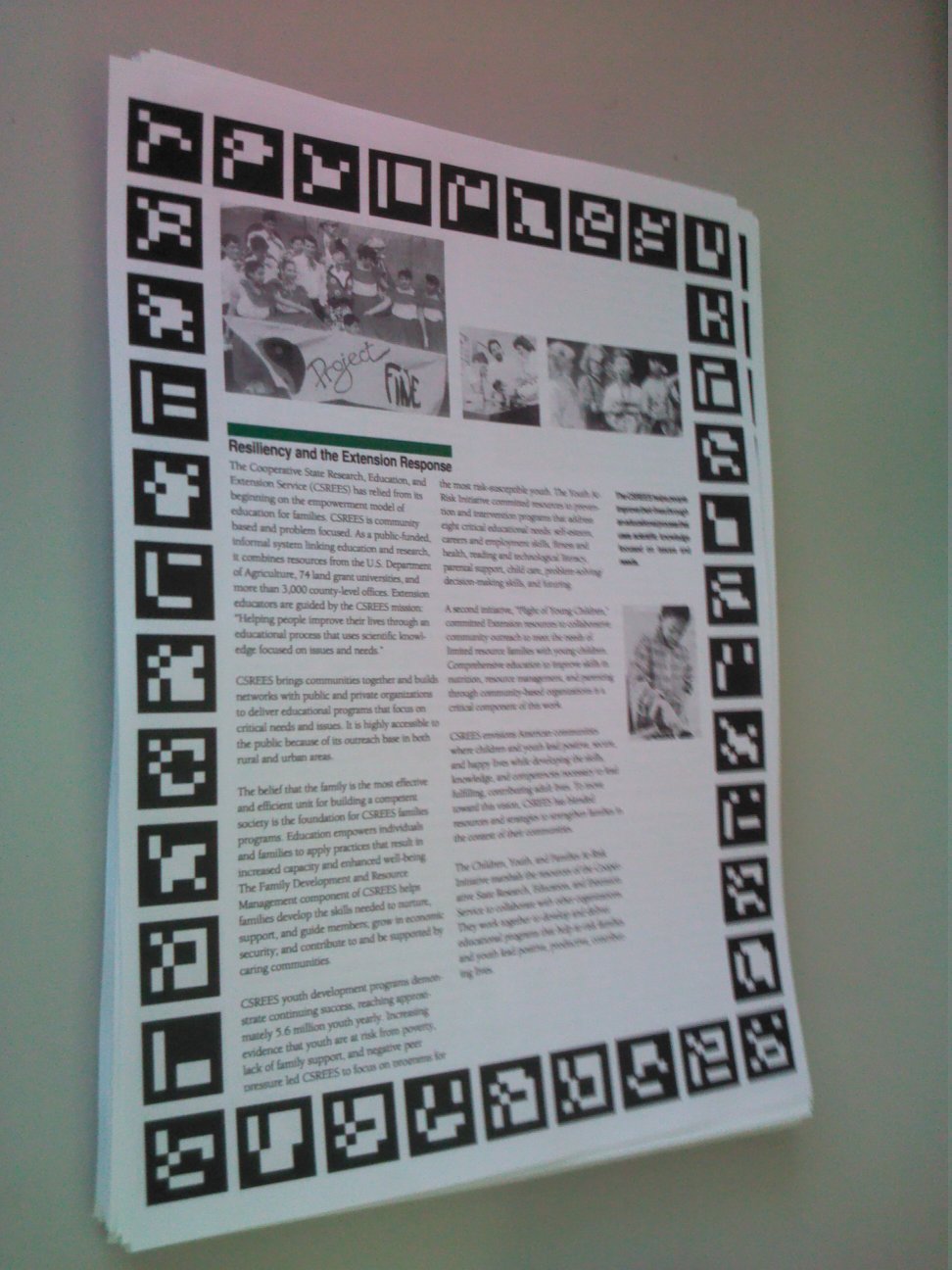}
      \end{subfigure}
    \caption{Representative examples of photographs in the B-MOD dataset.}
    \label{fig:photographs}
\end{figure}

From the collected templates, we obtained 19~725~photographs under unconstrained conditions.
About half of the photographs capture templates printed on A4 paper and half capture templates displayed on a 24 inch computer screen. 
Photographs of the templates were captured by multiple people instructed to use natural but challenging viewing conditions and to capture the pages preferably fast without much care for viewing angle and camera stability. 
This process resulted in a large variation in viewing angles and distances, and in diverse lighting  including shadows and non-uniform illumination. 
Many photographs also contain significant out of focus and motion blur. 
Representative examples of photographs are shown in Figure~\ref{fig:photographs}.

\begin{figure}[t]
  \centering
  \includegraphics{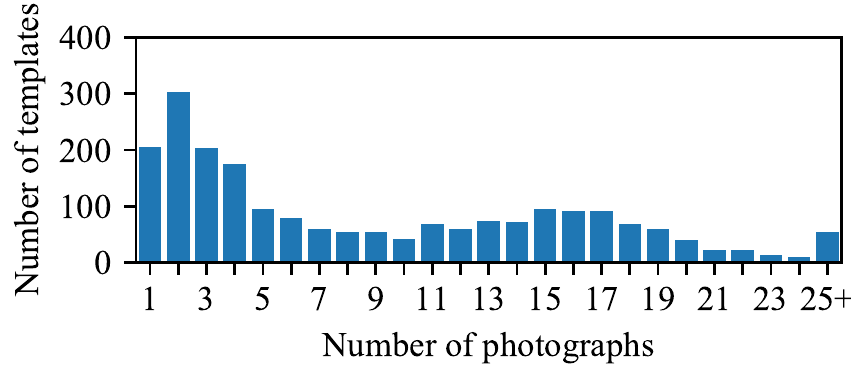}
  \caption{Histogram of the number of photographs taken per template.}
  \label{fig:histogram_photos}
\end{figure}

Altogether, 23~mobile devices were used to capture the photographs (smartphones and tablets). 
A detailed list of the used devices and the number of photos taken using them is shown in Figure~\ref{fig:devices} and a histogram of the number of photographs taken per page template is shown in Figure~\ref{fig:histogram_photos}.

\subsection{Page localization and alignment}
To localize the pages in photographs and to align them to the templates, we decided to add AR markers to the page edges as shown in Figure~\ref{fig:photographs}. 
As the combination of these markers is unique for each template and redundant, it robustly identifies it, thereby completely removing the need to separately track the photographed template identities. 
We localized the markers using the ARToolKit library\footnote{https://github.com/artoolkit} which is able to detect the markers even in strongly blurred images. 
In fact, the template identification failed only on images of such a poor quality that their analysis would be pointless anyway. 

Position of a template in a photo was estimated using RANSAC algorithm with homography transformation from all corner points of detected markers.
A homography captures projective transformation of a planar object caused by a perspective camera and it is able to explain large part of the observed geometric transformations; however, it is not perfect due to the uneven surface of the photographed pages and due to small geometric imperfections of the used cameras. 
We tried to remove these residual imperfections in alignment using non-rigid registration with only limited success as it often fails on strongly blurred images.
At the end, we used only the estimated homography which is precise enough to roughly align text lines in almost all cases.

\subsection{Line level ground truth}
In theory, it is possible to extract text information directly from PDF files.
However, such approach often fails, especially when some parts of the document are typeset as images or the document was created by scanning a physical document.
For that reason, we decided to extract text information from rendered template pages by processing them with two standard OCR products (ABBYY FineReader and Tesseract). 
We aligned both line transcriptions and we removed any lines where the two transcriptions differed. 
This process produced 64~221 unique text lines which represent 42~\% of all lines detected.
Line positions and heights were obtained solely using Tesseract and further refined.

We manually checked the correctness of the ground truth transcriptions on 300 randomly selected text lines containing 16~000 characters. From this sample, we estimated using bagging the maximum character error rate in the ground truth transcriptions to be at most 0.1~\% with 95~\% confidence.

\subsection{Text line position refinement}

\begin{figure*}[t]
  \centering
  \includegraphics[width=0.75\linewidth]{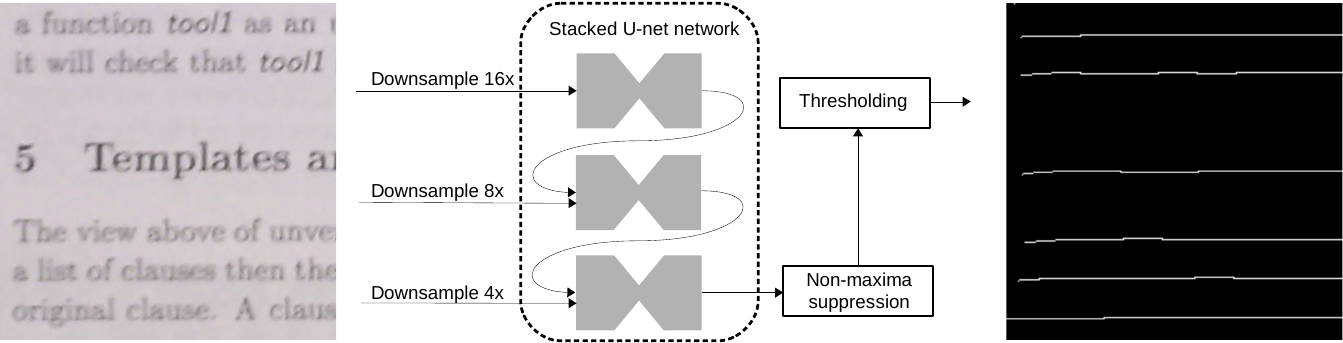}
  \caption{Overview of the baseline detection pipeline.}
  \label{fig:baselines}
\end{figure*}

Positions of the text lines mapped from the templates to the photographs using the estimated homography are sometimes not precise enough.
To correct these small miss-alignments, we opted to detect text baselines in homography-rectified photographs and to adjust the text line positions using these baselines.
We trained a baseline detection neural network on 25~manually annotated rectified photographs.

We decided to approach the baseline detection problem as a segmentation task aiming to identify pixels belonging to a baseline. 
We used the U-net architecture which has been previously shown to perform well in baseline detection~\cite{Diem2017}. 
We used a series of three stacked U-net modules with independent weights as shown in Figure~\ref{fig:baselines}.
Each of the modules produces pixel-wise baseline probability map. 
The three modules process the input image with progressively increasing resolution, each output  being $2\times$ upsampled and concatenated to the input of the following U-net module. 
The whole network was trained with a Dice loss end-to-end with the input images being down-scaled by factors of 16, 8 and 4 for the first, second and third U-net module, respectively. 
As the network produces slightly blurry probability maps, we apply vertical non-maxima suppression and subsequent thresholding to obtain binary baselines with single pixel thickness. 

We refine position of each text line by fitting a line using the RANSAC algorithm to the detected baseline points.  
This estimation uses baseline points contained in an initial axis-aligned rectangle of the text line moved downwards by one third of its height. 
This approach handles well cases when the baseline is only partially detected and it minimizes chance that the text line would be fitted to a neighboring baseline.
The text line bounding box is rotated and vertically translated to the position of the estimated baseline while restricting the maximum rotation angle.

\subsection{Ground truth verification}

Despite our best effort to precisely localize text lines in the photographs, two types of errors appear. 
Some text lines get mapped to a neighboring baseline, and positions of some text lines in extremely blurred images are simply poorly estimated. 
We tried to identify these miss-aligned text lines in order to remove them from the dataset.

We trained a text recognition neural network on all text lines (the network architecture is described in section \ref{sec:networks}). 
We removed 18~495~text lines on which the network achieved low character accuracy or high training loss normalized by ground truth transcription length. 
The two cut-off points were manually selected by visual inspection of sorted text lines. 

To solve the mismatch problem, we compared probability assigned by the trained network to the ground truth text and to the text decoded by the network.
We removed 2300 text lines for which the probability of the ground truth string was significantly lower compared to the probability of the decoded string.

\subsection{The final dataset}

\begin{figure*}
    \centering
    \begin{subfigure}{.3\linewidth}
        \centering
        \includegraphics[width=1\linewidth]{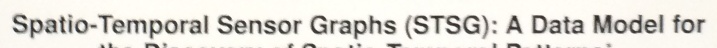}
        \includegraphics[width=1\linewidth]{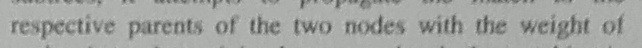}
        \includegraphics[width=1\linewidth]{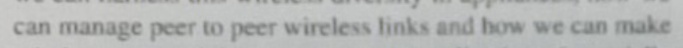}
        \includegraphics[width=1\linewidth]{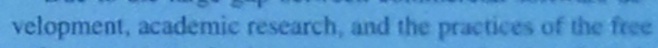}
        \includegraphics[width=1\linewidth]{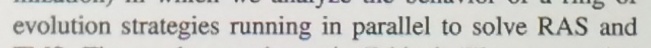}
        \caption{}
        \label{fig:line_easy}
      \end{subfigure}
      \hspace{2ex}
      \begin{subfigure}{.3\linewidth}
        \centering
        \includegraphics[width=1\linewidth]{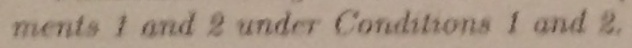}
        \includegraphics[width=1\linewidth]{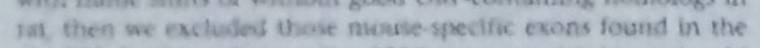}
        \includegraphics[width=1\linewidth]{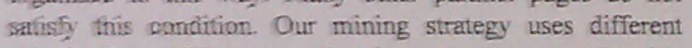}
        \includegraphics[width=1\linewidth]{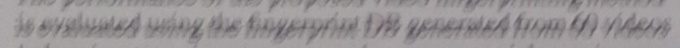}
        \includegraphics[width=1\linewidth]{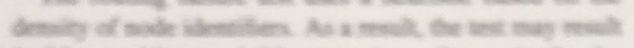}
        \caption{}
        \label{fig:line_medium}
      \end{subfigure}
      \hspace{2ex}
      \begin{subfigure}{.3\linewidth}
        \centering
        \includegraphics[width=1\linewidth]{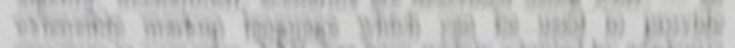}
        \includegraphics[width=1\linewidth]{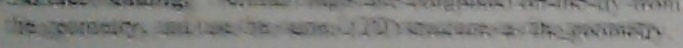}
        \includegraphics[width=1\linewidth]{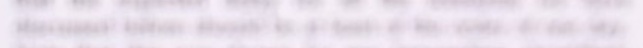}
        \includegraphics[width=1\linewidth]{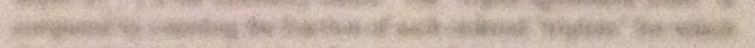}
        \includegraphics[width=1\linewidth]{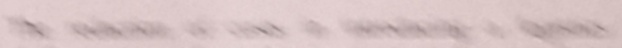}
        \caption{}
        \label{fig:line_hard}
      \end{subfigure}
    \caption{Samples of lines. Column (a) shows easy lines, (b) shows medium lines and hard lines are in (c).}
    \label{fig:samples_lines}
\end{figure*}

The dataset is available online\footnote{https://pero.fit.vutbr.cz/brno\_mobile\_ocr\_dataset}. 
It contains template pages rendered at 300 dpi, original photographs, rectified page photographs, PAGE XML~\cite{pletschacher2010page} ground truth files for both original photographs and rectified page photographs, and extracted line images normalized to height of 48 pixels.

In total, the final dataset consists of 515~400~cropped text lines, which are divided into 9~disjoint sets. 
The first division is into training, validation and test set. 
The second division is based on the difficulty of the text lines. 
We define 3 levels of difficulty: easy, medium and hard. 
The partitioning of the dataset is shown in Table \ref{tab:dataset}.
Histogram of extracted lines per template is shown in Figure \ref{fig:histogram_lines}.

\begin{figure}[t]
  \centering
  \includegraphics{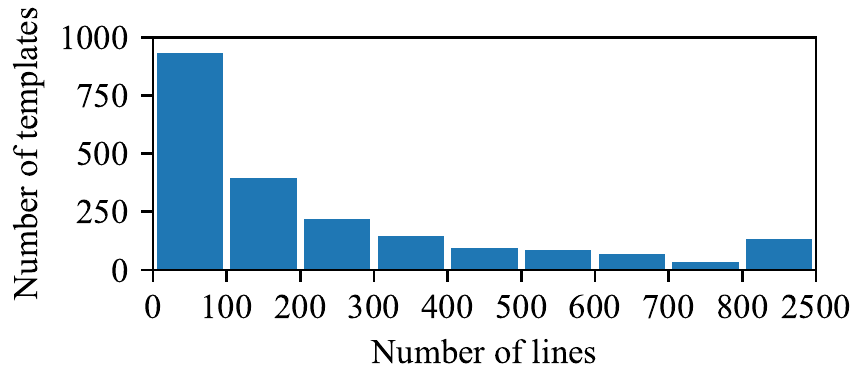}
  \caption{Lines extracted per template.}
  \label{fig:histogram_lines}
\end{figure}

\begin{figure}[t]
  \centering
  \includegraphics{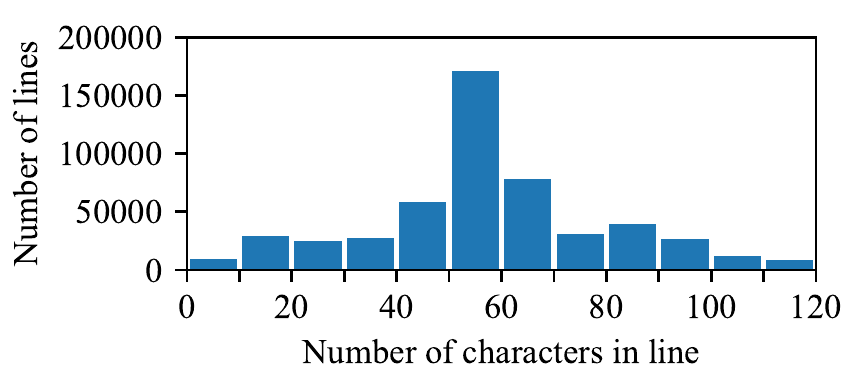}
  \caption{Histogram of the number of character in extracted lines.}
  \label{fig:histogram_lines}
\end{figure}

\begin{table}
	\vskip16pt
	\centering
    	
    	\begin{tabular}{ccccc}
            \toprule
            
            \multirow{2}{*}{
                \begin{tabular}[c]{@{}c@{}}\addlinespace[0.05cm] Difficulty\\ level\end{tabular}
                }
            
            & \multicolumn{3}{c}{Sets} & 
            
            \multirow{2}{*}{
                \begin{tabular}[c]{@{}c@{}}\addlinespace[0.05cm] Total by\\ difficulty level\end{tabular}
            }
            
            \\ \cline{2-4} \addlinespace[0.1cm]
            
            & Training & Validation & Test & \\ 
    		\midrule
            Easy   & 292 738      & 38 425         & 40 447   & 371 610 \\ 
            Medium & 95 239       & 11 891         & 16 411   & 123 541 \\
            Hard   & 15 767       & 1 775           & 2 707     & 20 249  \\
            Total by set  & 403 744      & 52 091         & 59 565   & 515 400 \\
    		\bottomrule
    	\end{tabular}
	
	\caption{Division of the resulting dataset. Values in the cells represent number of lines in the given set with certain difficulty level. }
	\label{tab:dataset}
\end{table}

The division into training, validation and test sets is based on the templates, which are divided into these subsets approximately by ratio $8:1:1$. 
As every template has different number of captured photographs, the number of photographs and lines in these subsets does not follow the original ratio, but on the other hand it ensures that the trained recognition system can't access text from the validation and test sets.

The second division is defined by training accuracy of a neural network trained on the whole dataset. 
A line is marked as easy, when the network transcribes the line without any errors.
If the training character error rate is below 20~\%, the line is assigned to medium difficulty. 
Otherwise, the line is marked as hard. 
Representative lines of each difficulty level are shown in Figure \ref{fig:samples_lines}.

\subsection{Text recognition evaluation protocol}

Measuring the text recognition accuracy of methods on our dataset is done using two metrics, character error rate and word error rate. 
Both of these metrics are computed on all three difficulty levels of the validation and test subsets as well as on the undivided subsets.

We compute character error rate as a sum of the number of errors at each transcribed line divided by the total length of the ground-truth transcription of the dataset part.
Similarly, word error rate is computed as a ratio between the number of error words and the total number of words in the set. 
The number of errors (characters or words) is obtained after line-level alignment which minimizes Levenshtein distance between the automatic transcription and ground truth.

The ground truth text for the testing part of the dataset is not public and an evaluation server is provided\footnote{https://pero.fit.vutbr.cz/brno\_mobile\_ocr\_dataset} to score results on the test set. 
The number of submissions is limited to discourage method optimization on the test set.

\section{Text recognition baseline}
\label{sec:networks}

This section describes neural network for text recognition that can be used as a baseline and which was used during processing and filtering of the dataset. In the first part we propose architecture of the network and in the second we compare achieved results.

\subsection{Baseline text recognition network}

End-to-end text line recognition networks can be build on three basic principles: as mixed convolutional and recurrent network with Connectionist Temporal Classification (CTC)~\cite{graves2006connectionist}, as a sequence to sequence model with recurrent encoder and an autoregressive recurrent decoder utilizing attention over the input~\cite{Kang2018}, or the recurrent parts of the encoder and decoder can be replaced by pure attention mechanism~\cite{Vaswani2017}. 
All of these architectures are able to capture both visual appearance as well as complex language dependencies.

We chose a network with CTC loss as a baseline for its simplicity and generally still competitive results.
The network starts with a 2D convolutional part inspired by the VGG networks~\cite{simonyan2014very}, which is followed by vertical aggregation using a convolutional kernel of single pixel width. Further layers of the network are 1D convolutions in the horizontal direction and bi-directional recurrent layers.
We use batch normalization and ReLU after each trainable layer.
We use Gated Recurrent Unit (GRU) layers~\cite{cho2014learning} in the recurrent part. 
Dropout regularization is applied in several points in the network. 
The network also contains several forward skip connections which improve propagation of information during network inference and of gradients during training. The trained network is available online\footnote{https://github.com/DCGM/B-MOD}.

\begin{table}
	\vskip16pt
	\centering
	
	\begin{tabular}{@{\extracolsep{4pt}}@{\kern\tabcolsep}crrrr}
        \toprule
        \multirow{2}{*}{\begin{tabular}[c]{@{}c@{}}\addlinespace[0.05cm] Difficulty\\ level\end{tabular}} & 
        \multicolumn{2}{c}{Test set} & 
        \multicolumn{2}{c}{Validation set}  \\ 
        \cline{2-3} \cline{4-5} \addlinespace[0.1cm]
        
         & \multicolumn{1}{c}{CER} & \multicolumn{1}{c}{WER} & \multicolumn{1}{c}{CER} & \multicolumn{1}{c}{WER} \\
         
		\midrule
        Easy           &   0.33 \%  &   1.93 \%  &   0.27 \%  &   1.58 \%  \\ 
        Medium         &   5.65 \%  &  22.39 \%  &   5.65 \%  &  21.88 \%  \\
        Hard           &  32.28 \%  &  72.63 \%  &  31.25 \%  &  71.71 \%  \\
        Total by set   &   3.15 \%  &  10.71 \%  &   2.48 \%  &   8.44 \%  \\
        \bottomrule
	\end{tabular}
	\vskip16pt
\begin{tabular}{@{\extracolsep{4pt}}@{\kern\tabcolsep}crrrr}
        \toprule
        \multirow{2}{*}{\begin{tabular}[c]{@{}c@{}}\addlinespace[0.05cm] Difficulty\\ level\end{tabular}} & 
        \multicolumn{2}{c}{Test set} & 
        \multicolumn{2}{c}{Validation set}  \\ 
        \cline{2-3} \cline{4-5} \addlinespace[0.1cm]
        
         & \multicolumn{1}{c}{CER} & \multicolumn{1}{c}{WER} & \multicolumn{1}{c}{CER} & \multicolumn{1}{c}{WER} \\
         
		\midrule
        Easy           &   0.50 \%  &   2.79 \%  &   0.40 \%  &   2.19 \%  \\ 
        Medium         &   7.82 \%  &  28.50 \%  &   7.79 \%  &  27.53 \%  \\
        Hard           &  39.76 \%  &  80.69 \%  &  38.70 \%  &  79.48 \%  \\
        Total by set   &   4.19 \%  &  13.39 \%  &   3.30 \%  &  10.45 \%  \\
        \bottomrule
	\end{tabular}
	
	\caption{The upper table shows results of the network with GRU layer, the second table shows results for the fully convolutional network. }
	\label{tab:results}
\end{table}

\subsection{Results}

To train the baseline networks we used all text lines from the training subset.
We used Adam optimizer with learning rate $2 \times 10^{-4}$ and batches of 32 text lines at most 2048 pixels wide. 
We trained two baseline networks: one fully convolutional without any recurrent layers, and one with recurrent layers as described in the previous section.
Overall results are shown in Table \ref{tab:results}.

\section{Conclusions}
\label{sec:conclusions}

We presented a new dataset for document Optical Character Recognition called Brno Mobile OCR Dataset (B-MOD). 
This dataset is mainly intended for development and evaluation of methods for document analysis and text recognition on challenging low-quality images.
In contrast to existing datasets, B-MOD is focused on mobile devices, and is diverse in terms of capturing devices, viewing angles, illumination conditions and image quality. 
The dataset provides precise line locations with associated highly accurate transcription ground truth.
We also propose a baseline text transcription neural network for this dataset which shows that this dataset is challenging for current text recognition methods.

We intend to extend the dataset by adding pages with handwritten annotations and highlighted sections, and by adding different types of documents such as magazines.
Also, it should be possible to remove the AR markers and rely purely on baseline detection and automatic text transcription to map text lines to their ground truth strings.

\bibliographystyle{IEEEtran}
\bibliography{ebib}

\end{document}